\documentclass[journal]{IEEEtran}
\usepackage{cite}

\ifCLASSINFOpdf
  \usepackage[pdftex]{graphicx}
\else
  \usepackage[dvips]{graphicx}
\fi
\usepackage{amsmath}
\usepackage{algorithm}
\usepackage{algorithmic}

\usepackage{url}

\usepackage{CJKutf8}
\usepackage{amssymb}
\usepackage{xcolor}

\usepackage{filecontents}

\begin{document}
%
\title{RODA: Reverse Operation based Data Augmentation for Solving Math Word Problems}
%
%
%
\author{Qianying Liu,\textsuperscript{\rm 1}
    Wenyu Guan, \textsuperscript{\rm 2}
    Sujian Li, \textsuperscript{\rm 2} \\
    Fei Cheng, \textsuperscript{\rm 1} 
    Daisuke Kawahara \textsuperscript{\rm 3} 
    and Sadao Kurohashi \textsuperscript{\rm 1}
    \thanks{\textsuperscript{\rm 1} Graduate School of Informatics, Kyoto University; ying@nlp.ist.i.kyoto-u.ac.jp;\{feicheng, kuro\}@i.kyoto-u.ac.jp; \\
    \textsuperscript{\rm 2} Key Laboratory of Computational Linguistics, MOE, Peking University; \{guanwy, lisujian\}@pku.edu.cn;\\
    \textsuperscript{\rm 3} School of Fundamental Science and Engineering, Waseda University\\
    dkw@waseda.jp}
    }
\markboth{Journal of \LaTeX\ Class Files,~Vol.~14, No.~8, August~2015}%
{Shell \MakeLowercase{\textit{et al.}}: Bare Demo of IEEEtran.cls for IEEE Journals}
%



\maketitle

\begin{abstract}
Automatically solving math word problems is a critical task in the field of natural language processing.
Recent models have reached their performance bottleneck and require more high-quality data for training. 
We propose a novel data augmentation method that reverses the mathematical logic of math word problems to produce new high-quality math problems
and introduce new knowledge points that can \textcolor{black}{benefit learning the mathematical reasoning logic}. We apply the augmented data on two SOTA math word problem solving models and compare our results with a strong data augmentation baseline.
Experimental results show the effectiveness of our approach\footnote{We release our code and data at https://github.com/yiyunya/RODA.}.
\end{abstract}


%
\IEEEpeerreviewmaketitle

\section{Introduction}

\IEEEPARstart{S}{olving} Math Word Problems (MWPs) is the task that infers a mathematical expression and the final answer from the natural language description of a math problem, which has been crucial due to its importance of numerically reasoning in natural language (NLP) processing ~\cite{dua2019drop,nie-etal-2018-operation}. 
Figure \ref{fig:example}  shows two examples which include math word problems and their corresponding solution equations and results.

\begin{figure*}
        \centering
      \begin{tabular}{p{195pt}|p{195pt}}
      \hline
         \hline
         MWP1 &   MWP2 \\
         \hline
        The distance between city A and B is \textbf{660 km}, the car starting from \textbf{A} drives \textbf{32 km/h}, and the car starting from \textbf{B} drives \textbf{34 km/h}. The two cars are starting from the two places \textbf{at the same time} \textbf{heading toward each other}. How many hours later would the two cars \textbf{meet}? & The car starting from A drives 32 km/h, and the car starting from B drives 34 km/h. The two cars are starting from the two places at the same time heading toward each other. \textit{10 hours later the two cars would meet.}\textit{ What is the distance between city A and B?}\\
        \hline
         Equation: $x = 660 / ( 32 + 34 )$ & Equation:  $x^\prime = 10 * (32 + 34)$\\
         \hline
        \textcolor{black}{Knowledge: Time = Distance / Speed}&\textcolor{black}{Knowledge: Distance = Time * Speed}\\
       \hline
        Answer: 10 &  
        Answer: 660\\
        \hline
        \hline
      \end{tabular}
      \caption{Two MWP Examples. The solution equations and \textcolor{black}{mathematical knowledge} of MWP1 and MWP2 are reversed to each other. 
      }
      \label{fig:example}
\end{figure*}
\textcolor{black}{Various manually constructed MWP datasets have advanced the development of MWP solving has been advanced, including Math23K~\cite{shi2015automatically} and Dolphin18k~\cite{shi2015automatically}.}
With these datasets as training data, the recent mainstream approaches are based on sequence-to-sequence (seq2seq) neural networks (NN)\textcolor{black}{,} where a math word problem serves as the input sequence and a math equation as the output sequence.
However, these approaches have reached their performance bottleneck as the size of the available data is still far from enough.
Thus, how to augment data is the \textcolor{black}{primary} consideration for improving the performance of MWP solving.

Data augmentation for NLP has been a critical and challenging research topic, especially in the case of MWP solving. 
Traditional data augmentation methods are mainly based on paraphrase generation, \textcolor{black}{such as EDA~\cite{wei-zou-2019-eda}, which randomly edits words, and back translation~\cite{DBLP:conf/iclr/YuDLZ00L18}, which translates the example to an auxiliary language and then back to the original language to paraphrase the example.} These methods have two \textcolor{black}{key} drawbacks when applied to MWPs.

First, due to the preciseness of mathematics, the text description of each math word problem must be absolutely rigorous, so that even missing only one keyword could make the information incomplete and the problem unsolvable. As shown in Figure \ref{fig:example}, 
all key information points of the problem marked in bold cover nearly \textcolor{black}{one-third} of the text content. With any of the key points missed, the problem would become meaningless.  
Therefore, in the task of MWP solving, \textcolor{black}{paraphrase-based} methods may potentially produce noise, mislead the model and degrade the performance.
Second, the \textcolor{black}{critical} difficulty of MWPs solving lies in linking mathematical knowledge points and reasoning logic with examples. \textcolor{black}{Here the mathematical knowledge points refer to formulae that can solve the MWP, such as $area_{circle}=pi*radius^2$, while the reasoning logic refers to the abstractive understanding of the relationship between these concepts, which is the desired ability of the MWP solving model.} \textcolor{black}{Paraphrase-based} methods only introduce lexicon level variance but \textcolor{black}{do} not generate new examples on the mathematical level. Therefore, the performance improvement of \textcolor{black}{paraphrase-based} data augmentation methods in the task of MWP solving is limited. 
Recently, a new research line of logic-based data augmentation uses \textcolor{black}{task-specific} logic to construct new examples for Natural Language Inference and Question Answering \cite{kang2018adventure, asai-hajishirzi-2020-logic, gokhale2020vqa}. These methods \textcolor{black}{introduce} reliable rule-based logic, which does not produce noise, to generate logic-level new examples. However, existing methods are task specifically designed and cannot be directly applied to MWP solving. Meanwhile, these methods are based on one or \textcolor{black}{two-step} simple logic operations and are relatively simple.

In this paper, unlike \textcolor{black}{the} previous practice of data augmentation,  \textcolor{black}{we instead simulate human double-checking and propose an MWP generation method to obtain more high-quality MWPs that are inferred through the reverse operation of the original problems.} 
As we observe,
when humans solve MWPs, a common technique to \textcolor{black}{guarantee accuracy} is to perform double-checking on the problem. As shown in Figure \ref{fig:example},  MWP1 asks about the time that the two cars have spent before meeting each other, and its solution implicitly holds the \textcolor{black}{mathematical knowledge point} that
${time = distance / speed}$.
With this \textcolor{black}{knowledge}, we can get the equation $x=660/(32+34)$, where $10$ is the final answer of $x$ and denotes the spent time.
To verify the correctness of the solution of MWP1, \textcolor{black}{humans conceive a reverse problem MWP2} with the \textcolor{black}{mathematical knowledge point} ${distance = time*speed}$, which takes the answer of MWP1 (i.e., $x=10$) as a known quantity and the distance $x^\prime$ as unknown.
\textcolor{black}{We can efficiently produce a new math word problem in the reverse operation and ensure its solution quality.}
Concretely, we can seek one known quantity (e.g., $660$ in MWP1) in the original problem and change its surrounding declarative description into an interrogative sentence (i.e., the last sentence in MWP2). At the same time, we change the original interrogative description about the unknown quantity  into a declarative statement with the original solution (e.g., $x=10$) substituted (i.e., the next to last sentence in MWP2). Then, with most content unchanged, we can obtain a new math problem (e.g., MWP2) from the original problem (e.g., MWP1).

We can see that this kind of reversion-based data augmentation has the following benefits: First, this way of generating new data is relatively simple and reliable so that the key information will not be lost; \textcolor{black}{Second, the reverse operation can infer new knowledge points which helps to learn mathematical reasoning logic;} Third, more high-quality data can be used \textcolor{black}{to train the neural networks well.}
Next, we combine this Reverse Operation based Data Augmentation (RODA) with seq2seq models
and conduct experiments on Math23K, \textcolor{black}{the most influential large-scale dataset for MWPs.} To be noticed, our method could be easily adapted to any supervised model on high-quality \textcolor{black}{arithmetic MWP corpora with equation solutions}.
Experimental results show that our method could benefit various models
and outperform previous state-of-the-art results on solving MWPs.

In summary, our contributions are three-fold:

\begin{itemize}
\setlength{\itemsep}{0pt}
\item We follow the double-checking mechanism and propose the reverse operation based data augmentation method, which is 
easy to apply and accurate.

\item \textcolor{black}{We show how our method can introduce new mathematical knowledge points to the new examples and help to learn the reasoning logic.}
\item We effectively use data augmentation results in the MWP solving and achieve the state-of-the-art on the Math23K dataset.
\end{itemize}

\section{Related Work}

\subsection{Math Word Problem Solving}

\subsubsection{Models}
Early approaches of solving MWPs mainly rely on predefined rules to map the problems into several predefined templates \cite{Bobrow:1964:NLI:889266,Charniak:1969:CSC:1624562.1624593}. Recent studies on solving MWPs adopt either semantic parsing techniques to parse the natural language questions into logic forms \cite{roy2015solving,roy2017unit,roy2018mapping,shi2015automatically,zou2019text2math}, or question answering style end-to-end models to directly predict the solution equations \cite{ling2017program,wang2019template}. \textcolor{black}{{Wang et al.}~\cite{wang2017deep}} first used seq2seq based models \textcolor{black}{to generate the mathematical solution of MWPs directly.} \textcolor{black}{{Wang et al.}~\cite{wang2018translating}} further extended the method by performing equation normalization as preprocessing and generating the suffix notation of the equation. \textcolor{black}{{Zhang et al.}~\cite{ijcai2020-0555}} further studied the diverse solution problem via multiple-decoders. Various studies \cite{liu2019tree, wang2019template, guan-etal-2019-improved,chiang2018semantically, DBLP:conf/ijcai/XieS19} further improved the model with \textcolor{black}{tree-structured information}. 
Attention mechanism and Graph Neural Networks have been studied to capture the intra-relation of the numbers~\cite{li2019modeling,zhang-etal-2020-graph}. 

\subsubsection{Datasets}


\begin{table}[ht]
\centering
\begin{tabular}{llll}
\hline

\bf Dataset & \bf \#Problem& \bf \#Template &\bf Equa Ans \\ \hline \hline
ALG514 & 514 & 28 & \checkmark \\
MAWPs & 2,373 & 344 & \checkmark \\
AllArith & 831 &- & \checkmark \\
\hline
Math23K & 23,162 & 3,527 & \checkmark\\
Dolphin18k & 18,460 &5,871 &  \checkmark\\
AQuA & 100,000&- & $\times$\\
MathQA &37,259 & -& $\times$\\

\hline
\end{tabular}
\caption{An Overview of Math Word Problem Datasets. The upper three datasets are early \textcolor{black}{datasets} that are relatively \textcolor{black}{small}. The bottom four datasets are recently constructed large scale datasets. \textbf{"Equa Ans"} stands for whether the final mathematical answer is expressed in mathematical equation style.}
\label{datasets}
\end{table}

Early datasets of MWP solving were designed for \textcolor{black}{rule-based} methods and are relatively small \cite{kushman2014learning, koncel-kedziorski-etal-2016-mawps, roy2015solving}. As shown in Table \ref{datasets}. These datasets range over a small variety of problem domains and answer equation templates. 

With the advent of data-driven methods especially neural networks, larger datasets are constructed. These datasets can be divided into two categories according to the answer format. 
In the first category, the mathematical solution is expressed as a mathematical equation. The most influential dataset is Math23k \cite{shi2015automatically}, \textcolor{black}{a Chinese dataset containing 23,162 elementary-school-level math word problems and equation solutions.} 
\textcolor{black}{The solution equation to each problem contains only one unknown variable.} 
Another large scale dataset is Dolphin18K \cite{shi2015automatically}, which is an English dataset that contains 18,460 \textcolor{black}{problems} along with the solutions. \textcolor{black}{The dataset contains multi-unknown variable problems, many minor typos, and wrong solutions. We do not use this dataset in our experiment that our data augmentation method pursuits are producing high-quality new data from high-quality data.} 
In the second category, the mathematical solution is expressed as semantic parsing style or descriptive text style output instead of equations, such as  {AQuA}~\cite{ling2017program} and {MathQA}~\cite{amini-etal-2019-mathqa}. Our method \textcolor{black}{does not apply to} these answer expressions since the new equation generation module could not be applied.
Here we conduct experiments on a small, simple English dataset AllArith and the large-scale, complex Chinese dataset Math23K to show how our method is effective among different languages and datasets.

\subsection{Data Augmentation}

The two most popular methods for \textcolor{black}{sentence-level} data augmentation in NLP \textcolor{black}{are} back translation \cite{DBLP:conf/iclr/YuDLZ00L18} and EDA \cite{wei-zou-2019-eda}. \textcolor{black}{{Yu et al.}~\cite{DBLP:conf/iclr/YuDLZ00L18}} used a \textcolor{black}{high-quality} machine translation system to translate the original text into a new language and then backward to perform paraphrase style data augmentation. \textcolor{black}{{Wei and Zou}~\cite{wei-zou-2019-eda}} slightly modified the input sentence on the token level by performing four kinds of operations: synonym replacement, random insertion, random swap, and random deletion. The drawback of applying these two methods on MWPs is that MWPs are very sensitive \textcolor{black}{to even slight modifications;} \textcolor{black}{any} key information missed the problem would become unsolvable. In addition, these methods only provide lexical level variance but no new reasoning knowledge. 

Recently due to the reliability of rule-based data augmentation in NLP, task-specific logic rules for data augmentation \textcolor{black}{have} been explored in Natural Language Inference (NLI) and Question Answering. \textcolor{black}{{Kang et al.}~\cite{kang2018adventure}} studied NLI-specific \textcolor{black}{logic-based data augmentation,} which generates new examples by replacing tokens or changing labels on the original training examples.
They only used the logic operations of NOT($\neg$) and \textcolor{black}{equivalence}($\Leftrightarrow$).
\textcolor{black}{{Asai and Hajishirzi}~\cite{asai-hajishirzi-2020-logic}} further extended entailment($\rightarrow$) operation for common domain  question answering data augmentation by measuring the transitive consistency of pairs of questions. \textcolor{black}{{Gokhale et al.}~\cite{gokhale2020vqa}} studied disjunction($\vee$) and conjunction($\wedge$) operation for yes-no style visual question answering.
All these studies involve only one or two steps \textcolor{black}{of} simple logic. By contrast, our method \textcolor{black}{uses} reversed operation of complex mathematical computation and can introduce new reasoning logic in the generated new examples.






\section{Methodology}
Given a high-quality MWP dataset, we propose RODA\textcolor{black}{,
producing} new accurate math word problems to enlarge the data scale. There is a linear correlation between the number of new questions generated and the number of known variables in the question. \textcolor{black}{Next,} we can use the augmented dataset to improve the supervised MWP solving models.



\subsection{Reversion  based Data Augmentation}

\textcolor{black}{To perform} data augmentation, we reverse the original problems in the dataset to new problems, and the reversion process  consists of three steps: number identification, problem transformation, and equation generation.
To ensure the quality of the new data, the main criteria of our reversion process is ``quality first quantity second'', so that our method relies on some well-designed empirical rules 
in the three steps.


\subsubsection*{Number Identification}
\label{Candidate}

\textcolor{black}{This stage consists of two steps. We first use an LSTM classifier to perform significant number identification and determine irrelevant numbers. Then we further filter out numbers that cannot perform reversed-based operation and then use exact match to map the numbers in the equation to the numbers in the question text. The statistics of this stage are given in Table \ref{tab:aug}}.

A MWP might contain various numbers that are irrelevant to the solution, such as the date or description text such as `tenth grade student'. \textcolor{black}{Following {Wang et al.}~\cite{wang2017deep}'s work, we perform significant number identification by building an LSTM-based classifier to determine the significance of the numbers and filtering out the irrelevant numbers. The classifier uses single layer LSTMs with 128 nodes and a symmetric window of length 3. The classification performance can reach around 99\% accuracy.} 

In addition, the numbers in a math word problem do not necessarily map  one to one with the numbers in the corresponding solution equation. \textcolor{black}{One} number may appear in the solution equation but do not appear in the word problem, and vice versa.
Thus, to conduct high-quality word problem reversion, we first need to identify the valid numbers which can be converted to an unknown quantity using the reverse logic of the original solution. 
Here, we propose a rule-based method to identify the possible numbers for reversion.
Four key rules are explained as follows, and we also show their examples in Table \ref{tab:validnum}.

\begin{table*}[t]
\centering
\begin{tabular}{p{40pt}|p{280pt}|p{25pt}|p{65pt}}
\hline
\textbf{Rule} & \textbf{MWP examples} & \textbf{Num} & \textbf{Equation}\\
\hline
\hline
\small{1} & \small{A has 4 piles of 2 apples and B has 2 apples. B gave 1 apple to A, how many does A have in total now?} & \small{2}& \small{$x = 4 * 2 + 1$}\\
\hline
\small{2} & \small{The side length of a square is 2, what is the area?} &\small{2} & \small{$x = 2*2$}\\
\hline
\small{3} & \small{The side length of a cube is 4, what is the volume?} & \small{4 \& 3}& \small{$x=4^3$}\\
\hline
\small{4} & \small{The diameter of a circle is 5, what is the perimeter?} &\small{$\pi$}& \small{$x=\pi*5$}\\
\hline
\end{tabular}
\caption{Examples of In-valid Numbers.}
\label{tab:validnum}
\end{table*}

\begin{enumerate}
\item \textbf{Problem Number Duplication}
If a number appears more than once in the problem, we filter this number out because we cannot map the numbers in the text to the equation with an injective function. 
As shown in Table \ref{tab:validnum}, 
we cannot know whether a separate `2' is related to \textit{A} or \textit{B}, and thus it is difficult for us to conduct a precise reversion.
\item \textbf{Equation Number Duplication}
If a number appears more than once in the equation, we filter this number out because it may not be capable \textcolor{black}{of being solved} with a linear equation with one unknown variable. To solve \textcolor{black}{higher-order} polynomial functions, introducing new operators such as root operation would be essential. \textcolor{black}{However,} such operators are out-of-domain (OOD) with the original data and would introduce noise.
\item \textbf{Power Operation}
If a number is involved with power operation, we filter it out because the inverse operation, which is logarithmic operation or root operation, is OOD. We filter out both the base number and the exponent number.
\item \textbf{Constant Term Numbers}
Constant term numbers are not applicable for reverse operation, such as $\pi$ and unit conversion terms.
\end{enumerate}

\textcolor{black}{We then perform exact match between the numbers in the question and the equation to align the numbers.}

\subsubsection*{Problem Transformation}
\label{QuestionCon}
After collecting a set of valid number candidates, we perform problem reversion for each number to get a new transformed math word problem.
The main work is to convert the original question sentence into a declarative sentence with a definite quantity and convert
the sentence with the identified number into a question sentence 
with its question point on the number.
Specifically, for the first conversion, we name the original question sentence as $Q$. \textcolor{black}{From $Q$,} we find the interrogative pronoun according to a list \textcolor{black}{compiled in advance and replace the pronoun with the answer of the original problem.} \textcolor{black}{Next,} we adjust the word order to make the sentence fluent and natural 
and get the declarative sentence $D^\prime$.


\begin{CJK*}{UTF8}{gbsn}
For the second conversion, given the candidate number $c_i$, we get the sentence $D$\textcolor{black}{,} which contains $c_i$. \textcolor{black}{Next,} we take $c_i$ as the question focus and change $D$ into a question sentence $Q^\prime$. For different languages, the conversion process is different. For example, for Chinese, the conversion is relatively simple and just replaces $c_i$ with an interrogative pronoun such as 
``多少 (How many)'' without the need of adjusting word order. \textcolor{black}{We show further details of the interrogative pronouns in the appendices.}

\end{CJK*}


For English problem transformation, we follow the manually encoded transformation rules from \textcolor{black}{{Heilman et al.}~\cite{heilman2009question}}. We extend the interrogative pronoun candidates to the following list: \textit{how many, how much, how far, how tall, how long, how fast, how old, how big, what fraction, what.}

\textcolor{black}{We then edit the original math word problem.} We delete the two sentences $D$ and $Q$, and add $D^\prime$ and $Q^\prime$ at the end of the text. Then we get the new transformed math word problem.
Our problem transformation method is simple but effective, which is the basis of producing new high-quality MWPs.

\subsubsection*{Equation Generation}
For each new transformed word problem, we need to generate its solution equation.
Similar to problem transformation, \textcolor{black}{we derive the new solution equation according to the solution equation of the original problem.}

\textcolor{black}{To make the process of generating the new equation clear,} we take the two MWPs in Figure \ref{fig:example} as examples. 
For the pre-identified number $660$ in MWP1, it will be changed into a variable (i.e., $x^\prime$). At the same time, we substitute the original variable $x$ with its answer $10$. Then we can get an intermediate equation $10=x^\prime/(32+34)$.


\begin{figure}[!h]
    \centering
    \includegraphics[scale=0.30]{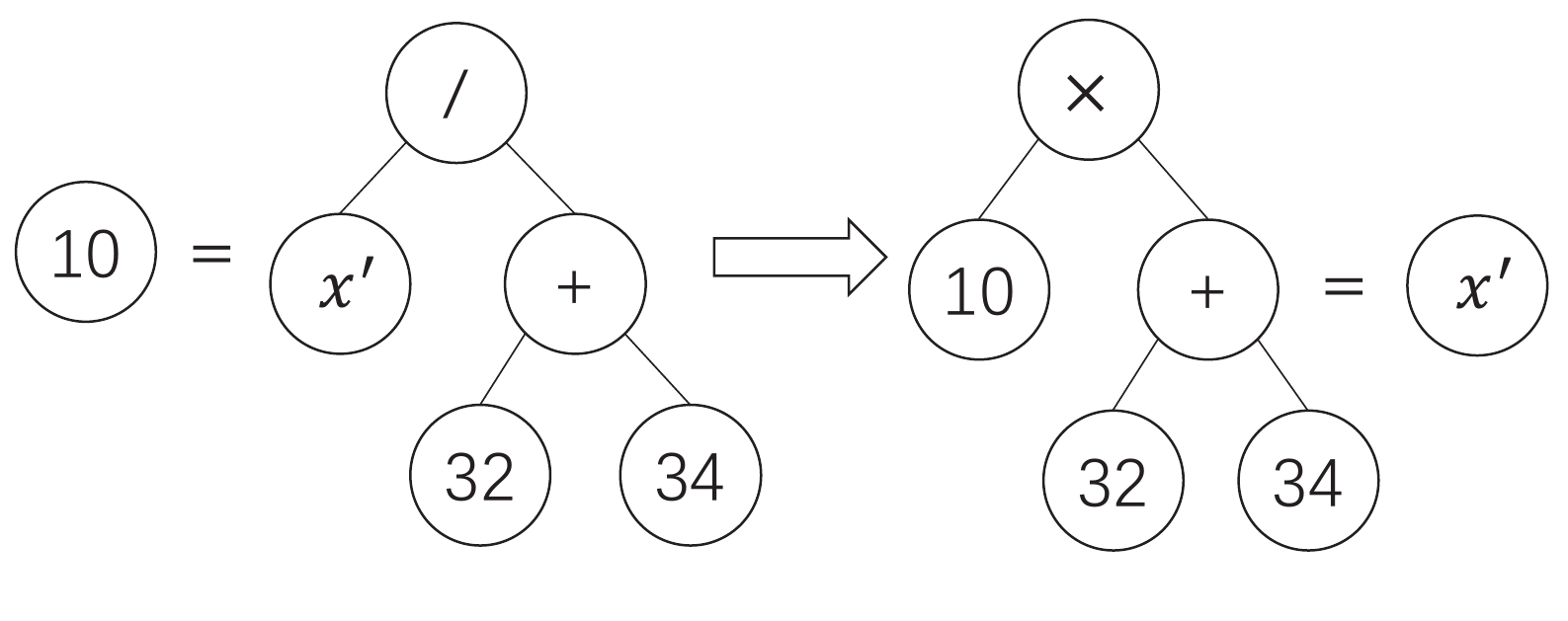}
    \caption{Example of equation conversion.}
    \label{fig:inverseExp}
\end{figure}

\begin{figure}[!h]
    \centering
    \includegraphics[scale=0.4]{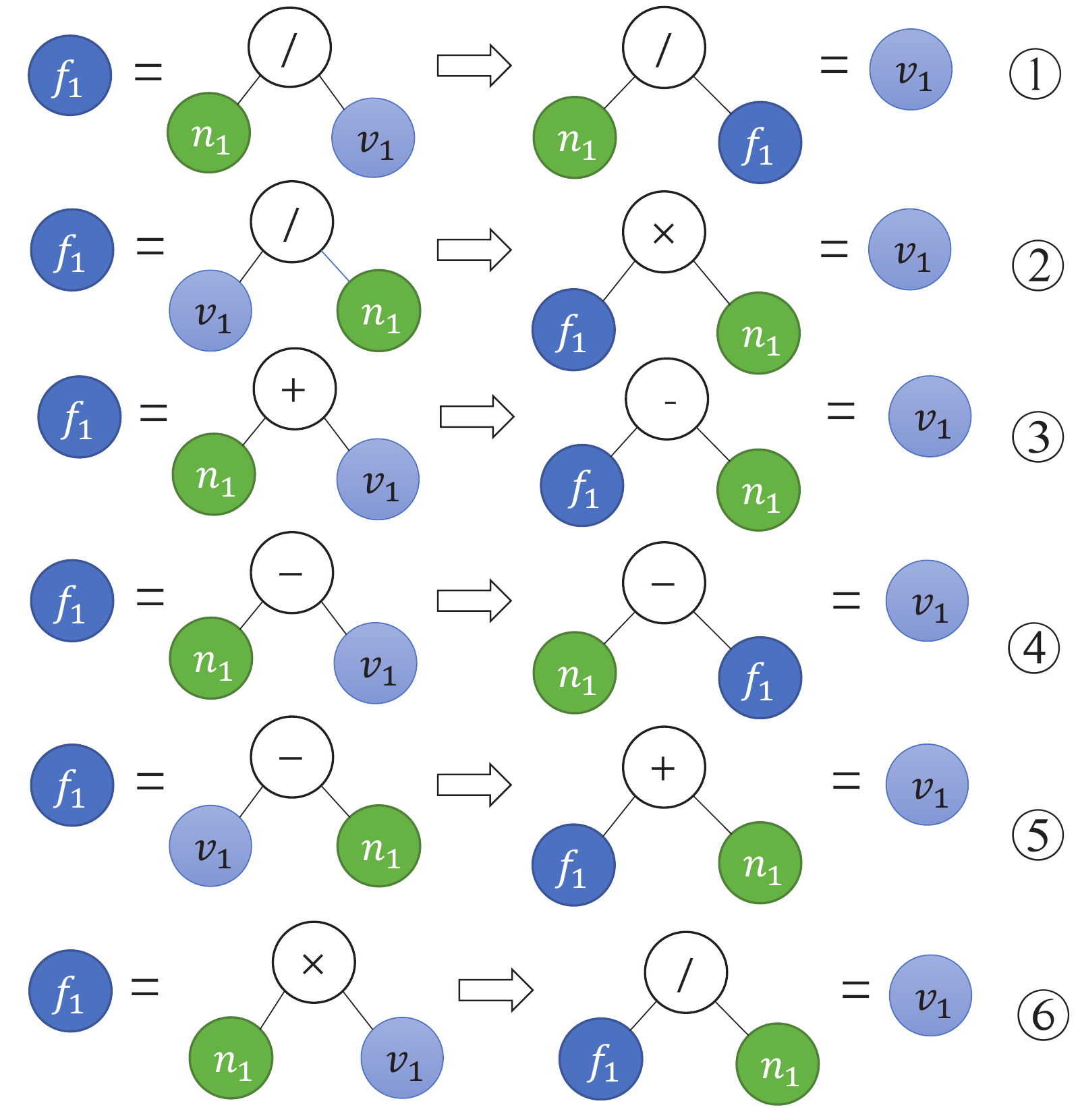}
    \caption{Rules of equation reversion}
    \label{fig:rule}
\end{figure}

\begin{figure}[!h]
\centering 
    \includegraphics[scale=0.25]{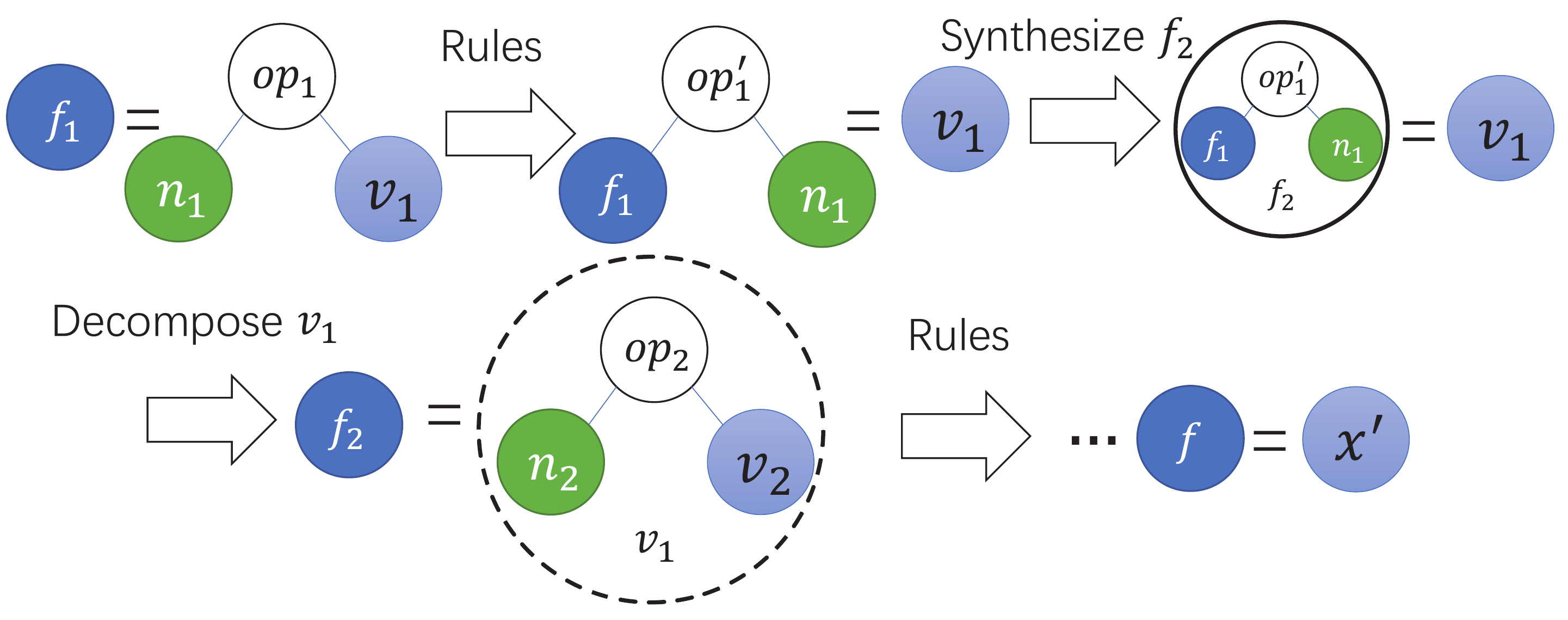}
    \caption{Illustration of recursive equation conversion.}
    \label{inveAbs}
\end{figure}

\begin{algorithm}
\caption{The algorithm of equation conversion}
\begin{algorithmic}[1]
\floatname{algorithm}{Procedure}
\renewcommand{\algorithmicrequire}{\textbf{Input:}}
\renewcommand{\algorithmicensure}{\textbf{Output:}}
\REQUIRE Left part of medium equation $l\_tr$ and right part of equation $r\_tr$ 
\ENSURE Inverse equation in tree structure
\STATE $root = r\_tr.root$\;
\STATE $num\_tr, var\_tr = find\_var(r\_tr)$\;
\WHILE{$root != x^{'}$}
\STATE$l\_tree = rule(l\_tr, num\_tr, root)$\;
\STATE$r\_tr = var\_tr$\;
\STATE$root = r\_tr.root$\;
\STATE$num\_tr, var\_tr = find\_var(r\_tr)$
\ENDWHILE
\RETURN $l\_tr = x^{'}$
\end{algorithmic}
\label{alg}
\end{algorithm}


\textcolor{black}{Next,} we need to convert this equation to its equivalent format where the variable $x^\prime$ is located on the right side of the equal-sign alone.
Figure \ref{fig:inverseExp} displays the equation conversion result of our running example.
To conduct equation conversion, we design a recursive conversion algorithm based on the syntax tree structure.
We first construct a quasi-binary syntactic tree for the original math equation.
We denote the part \textcolor{black}{on the left} to the equal sign as $f_1$(the formula in stage 1). For the part right to the equal sign, we build  a binary tree with one operator $op_1$ as the root node \textcolor{black}{with two child trees.}
The child tree which has the new variable $x^\prime$ is marked as $v_1$(the child tree with \textcolor{black}{a variable} in stage 1) and the other one as $n_1$(the child tree without \textcolor{black}{any variables} in stage 1), and this identifying process is named as function \textit{find\_var} in Algorithm 1. 
This step corresponds to the upper left part of Figure \ref{inveAbs}.
Next, like the upper middle part of Figure \ref{inveAbs}, we move $n_1$ to the left side and get a new operator $op_{1}'$ according to equation reversion rules, which are summarized from the basic mathematical computation.
We show thee rules in Figure \ref{fig:rule}.
Regarding $f_1$ and $n_1$ as two new child trees and $op_{1}'$ as the new root node, we can get a new tree $f_2$, which is the black circular ring in the upper right part of Figure \ref{inveAbs}, and we just use a single node $f_2$ to represent it in the following step. 


\textcolor{black}{Then, $v_1$ is further broken down} as a binary tree which is composed of the root node $op_2$ and the child trees $n_2$ and $v_2$ as shown in the \textcolor{black}{lower-left} part of Figure \ref{inveAbs}.
If we ignore the dotted circular sign of $v_1$, \textcolor{black}{in this state,} the equation has the same structure as the beginning state, so the following process will repeat until  $v_n$ has only one node $x^\prime$.
At this time, referring to the lower right part of Figure \ref{inveAbs}, all nodes of numbers and operators are moved to the left side and only the variable node $x^\prime$ is left on the right side.
The concrete process of equation conversion is shown in Algorithm \ref{alg}.


Since one math equation can be written in several equivalent forms (e.g., $4 + 2 -3$ and $4 - 3 + 2$), which \textcolor{black}{brings} noise to the model training, 
we conduct equation normalization for all the original and generated equations. We follow the criteria of \textcolor{black}{{Wang et al.}\cite{wang2018translating}} that if one equation could be converted into a shorter one, then it should be shortened, and the order of the numbers in one equation should follow their occurrence order in the problem text as much as possible. \textcolor{black}{{Wang et al.}~\cite{wang2018translating}} performed equation normalization by applying a series of rules that only work on limited cases such as a sequence of multiplication. Here we increase the applicable scope. We also use the simplification algorithm supported by sympy \cite{10.7717/peerj-cs.103} that heuristically simplifies the equations and \textcolor{black}{matches} with human writing regulations. \textcolor{black}{We show further details of the algorithm in the appendices.}

This equation normalization is essential for assuring the model performance since the equations generated by the reverse operation are written in a completely different style from that of the equations in the original data.  \textcolor{black}{The normalization can avoid domain shift of the prediction target equations.}

\subsection{MWP Solving Model}

Our data augmentation method can be combined with any preferred neural model.
Here, we adopt two seq2seq models and one classification model, then examine their performance.
First, we implement {Liu at el.}~\cite{liu2019tree}'s prefix model\textcolor{black}{,} which is a light and effective baseline.
At the same time, to show whether an advanced model can be further improved by our augmented data, we  use the SOTA model named Goal-driven tree-structured MWP solver (GTS) 
\cite{DBLP:conf/ijcai/XieS19}, which is an extension of the prefix baseline model. \textcolor{black}{We also build a classification model based on Transfromers following Kushman et al.~\cite{kushman2014learning} and Robaidek et al.~\cite{robaidek2018data}, which considers the equation template as the label of one MWP example. Here, we briefly introduce the three models.}

For the prefix model, formally, the model takes a sequence of tokens $\{x_i\}_{i=0}^{n}$ as the input and embeds them into a sequence of word embedding representations $\{e_i\}_{i=0}^{n}$ which are fed into a bidirectional long short term memory network (BiLSTM) encoder. 
Then two context-aware representations $h_{i}^{enc_f}$ and $h_{i}^{enc_b}$ are calculated and concatenated as $h_i^{enc}$ for each token.
Then these representations are given to the decoder to decode the output equation.


The decoder adopts a unidirectional LSTM to generate the output in an autoregressive manner. At each decoding time step $t$,
the decoder calculates the attention weight distribution $\{a_t^i\}$ on $\{h_i^{enc}\}$ with the embedding $e_t^{dec}$ of the output of the previous time step $y_{t-1}$ and the current hidden state $h_t^{dec}$ of the decoder LSTM, and assigns them to the encoder outputs $\{h_i^{enc}\}$ to form an attention-aware representation $s_t$ which is finally fed to a Multi-layer Perceptron (MLP) layer to generate the output token $y_t$.
\begin{align}
    h_t^d & = LSTM(h_{t-1}^{dec}, e_t^{dec}) \\
    \begin{split}
        s_t & = \sum_{i=1}^n \alpha_t^i \cdot h_i^{enc} \\
            & = \sum_{i=1}^n\frac{exp(h_i^{enc} \cdot h_t^{dec})}{\sum_{j=1}^n exp(h_j^{enc} \cdot h_t^{dec})} \cdot h_i^{enc}
    \end{split}
\end{align}



The GTS model \cite{DBLP:conf/ijcai/XieS19} further extends the prefix baseline with  subtree representations which can provide more information for the decoding process.
A recursive neural network is used to encode 
subtrees of the equation in a bottom-up manner. The subtree representation 
of one token $y_t$ is calculated based on its children nodes with the gate mechanism. With the subtree representations, this model can also well use the information of the generated tokens to predict a new token.

It is noted, for the seq2seq models, directly generating the solution equation with numbers \textcolor{black}{suffers} from a serious out-of-vocabulary (OOV) problem since the vocabulary of numbers is enormous. To address this problem, we follow \textcolor{black}{{Kushman et al.}~\cite{kushman2014learning}}, which used equation templates instead of actual equations as the prediction target of the model. The numbers in one MWP are notated as $temp_i$, where $i$ denotes the order \textcolor{black}{of} the numbers \textcolor{black}{that appear} in the problem. Extra constant numbers such as $\pi$ and $1$ are also added to the decoder vocabulary. Then the OOV problem can be solved. 


\textcolor{black}{For the classification model, we follow {Kushman et al.}~\cite{kushman2014learning} and {Robaidek et al.}~\cite{robaidek2018data}, which encodes the question text and then classifies the corresponding equation template. We add a [CLS] token to the question text and use a Transformer to encode the question text. We feed the feature vector of the [CLS] position to a multi-layer perception network to classify the equation template.}

To further improve the models, we mix and shuffle the augmented data with the original data as the new training set of the models. \textcolor{black}{It is noticed that our method does not limit to these three models.}

\section{Experiments}

\subsection{Experiment Setup}
In our experiments, we mainly experiment on Math23K\footnote{\textcolor{black}{Download link: https://github.com/SumbeeLei/Math\_EN}} which is the most influential large scale  dataset for MWP solving in the Chinese language. \textcolor{black}{It contains 23,162 one unknown variable elementary school level MWPs with the corresponding equation solutions. In train-test setting, the dataset is splitted to 21,162 training examples, 1,000 validation examples and 1,000 test examples.}


\textcolor{black}{For the classification model, we use Transformer base with 12 layer, 768 hidden state size and 12 attention heads. The classification model has 5693 labels. We train with a 16 batch size for 20 epoches. We use the SGD optimizer with a learning rate of 2e-5 and weight decay of 0.01. }

For the two seq2seq MWP solving models, we fix their embedding size to be 128.
For the prefix baseline model, the dimension of encoder hidden state is set to 512 while the dimension of decoder hidden state is 1024.
For the GTS model, the hidden state of both encoder and decoder is 512.
We use Adam optimizer to optimize these parameters.
The batch size is 128.
\textcolor{black}{To compare the performance with baselines fairly,
GTS model is tested by both train-test setting same as the other models, and also 5-fold cross validation setting same as the original paper}\footnote{We filter out the augmented samples of test set for each cross. So in the training stage, models can only learn from the training set and their augmented samples.}, while others are tested on the test set\footnote{We use the same split as {Wang et al.}~\cite{wang2018translating}. Only training set is used for augmentation.}. The experiments are done on GTX 1080Ti GPU, with a runtime of 10 hours for prefix baseline, 110 hours for GTS and \textcolor{black}{15 hours for the classification model.}

For data augmentation, we perform \textcolor{black}{the} reverse operation on the training set of Math23K\textcolor{black}{,} which includes 21,162 MWPs. From these problems\textcolor{black}{,} we can get 58,699 numbers \textcolor{black}{by} using the LSTM-based classifier.
At the same time, we filter out 1,490 problems which are composed of only numbers and operators \textcolor{black}{such as 'Please calculate 5+7*10'}, because they can not give \textcolor{black}{effective} supervision to the MWP solving models.
We also exclude 7,399 numbers in the problems which are not easy to be reversed, because they do not map one to one with the numbers in the solution equations as stated in Section 3.1. Finally
we totally get  47,318 new problem-equation pairs.
We can see, our data augmentation method successfully generates \textcolor{black}{new data whose size is more than two times the original data}, demonstrating the method's ability for performing large-scale data augmentation cheaply. We illustrate the statistics of the data augmentation results in Table \ref{tab:aug}.

\textcolor{black}{We also show the statistics of template coverage on the development set of the original training set and the augmented training set. As we can see our data augmentation method increased the template coverage of the development set, reducing the rate of uncovered templates by 21.5\%. This can show how our method introduces new mathematical knowledge points and benefit the model.}


\begin{table}[t]
\centering
\begin{tabular}{l|ll}
\hline
Type & \# & Prop.\\
\hline
Original Problems & 21,162 & -\\
Filtered Problems& 1,490 & 0.07\\
Original Numbers & 58,699 & 2.77\\
Candidate Numbers & 54,717 & 2.59\\
Irreversible Numbers & 7,399 & 0.35\\
Augmented Problems & 47,318 & 2.24\\
\hline
\end{tabular}
\caption{The statistics of the data augmentation on the training set of Math23K. \textbf{"Prop."} stands for the proportion of the item compared with original problems. 
}
\label{tab:aug}
\end{table}

\begin{table}[t]
\centering
\color{black}
\begin{tabular}{ll|ll}
\hline
Type& & \# & Prop.\\
\hline
\multicolumn{2}{l|}{Dev Set Templates} & 377 & -\\
\multicolumn{2}{l|}{Original Covered} & 307 & 81.4\\
& + RODA & 322 & 85.4\\
\hline
\end{tabular}
\caption{The statistics of the template coverage of Math23K train set templates on development set.  \textbf{"Prop."} stands for the proportion of the item compared with original problems. 
}
\label{tab:aug}
\end{table}

\subsection{MWP Solving Results}

\begin{table}[ht]
\centering
\begin{tabular}{lll}
\hline

\bf Model &  & \bf Acc \\ \hline \hline
Retrieval & Cosine \cite{robaidek2018data} & 23.8\% \\
& Jaccard \cite{robaidek2018data} & 47.2\% \\\hline
Classification & Transformer \cite{robaidek2018data} & 56.8\% \\
& Bi-LSTM \cite{robaidek2018data} & 57.9\% \\\hline
Generation
& DNS \cite{wang2017deep} & 58.1\% \\
& BiLSTM+Suffix+EN \cite{wang2018translating} & 66.7\%\\
& TreeLSTM\cite{liu2019tree} & 69.0\%\\
& Group-Attention\cite{li2019modeling} & 69.5\%\\
\hline
Ensemble & DNS+Retrieval \cite{wang2017deep} & 64.7\% \\
& DNS+suffix+EN Ensemble \cite{wang2018translating} & 68.4\% \\
\hline
\hline
\textcolor{black}{Classification}&\textcolor{black}{Transformer Ours} & \textcolor{black}{54.9\%} \\
&\textcolor{black}{Transformer Ours+RODA} & \textcolor{black}{63.7\%} \\
\hline
Generation&Prefix\cite{liu2019tree} & 67.8\%\\
&Prefix+RODA & \textbf{70.5}\%\\
\cline{2-3}
&\textcolor{black}{GTS}\cite{DBLP:conf/ijcai/XieS19} & \textcolor{black}{75.6\%}\\ 
&\textcolor{black}{GTS+RODA}&\textcolor{black}{\textbf{77.9\%}}\\
&GTS$\dagger$\cite{DBLP:conf/ijcai/XieS19} & 74.3\%\\ 
&GTS+RODA$\dagger$&\textbf{76.0\%}\\
\hline
\end{tabular}
\caption{Math word problem solving accuracy on Math23K. $\dagger$ denotes that the result is 5-fold cross validation performance. All other models are tested on the test set.}
\label{result}
\end{table}

We first evaluate the MWP solving performance by comparing our methods with other baseline methods. Here we name our Reverse Operation based Data Augmentation method RODA. In this experiment, we use all the 47,318 augmented data for training the prefix and GTS models.
Table \ref{result} shows the results of our methods and other novel systems on the Math23k \textcolor{black}{evaluated} by the final answer accuracy.
In this table, we classify the MWP solving models as retrieval-based, classification-based, generation-based and ensemble models.
The retrieval-based models mainly \textcolor{black}{calculate} a similarity score for questions in the test set and the questions in the training set and assign the template that has the highest similarity \cite{upadhyay2017annotating, robaidek2018data}. \textbf{Cosine} and \textbf{Jaccard} respectively denote the methods which adopt the corresponding similarities.
The classification-based models train a classifier to predict an equation template for each problem in a multi-class classification manner \cite{kushman2014learning}. 
For retrieval and classification models, we use the results from \textcolor{black}{{Robaidek et al.}~\cite{robaidek2018data}}. 

The generation-based models are the recent mainstream for MWP solving and  use end-to-end seq2seq models to directly generate an equation template. {Wang et al.}~\cite{wang2017deep} proposed the \textbf{DNS} model, which used seq2seq with significant number identification to generate an equation template. {Wang et al.}~\cite{wang2018translating} improved their model and proposed the \textbf{Suf+EN} model, which extends the DNS model by decoding the suffix notation and performs equation normalization for preprocessing. \textbf{TreeLSTM} \cite{liu2019tree} uses a top-down tree-structured decoder to predict the equations. \textbf{Group-Attention} \cite{li2019modeling} uses various attention methods to capture the intra-relation of the numbers. 
We also use two ensemble models \textbf{DNS+Retrieval} and \textbf{Suf+EN Ensemeble} for comparison, which \textcolor{black}{uses} bagging to combine the results of different models.

Our data augmentation method is exerted on two generation models Prefix \cite{liu2019tree} and GTS \cite{DBLP:conf/ijcai/XieS19} which have been introduced in Subsection 3.2. We choose these models as they can somewhat be representative of generation-based methods\textcolor{black}{,} especially GTS achieves the SOTA performance.
From Table \ref{result}, we can see that generation-based methods generally outperform retrieval-based and classification-based methods. \textcolor{black}{To show the efficiency of our method, we conduct experiments on both the classification models and the SOTA generation models. We can see that our method gains 8.8 points of improvement over our Transformer baseline classification model}. 
\textcolor{black}{We also show that our RODA method can further promote the current SOTA models:
Prefix+RODA and GTS+RODA boost Prefix and GTS by about 2.7 points and 2.3 points, respectively. Under 5-cross validation setting, our model also improves GTS by 1.7 points.} This also exhibits that more high-quality data is useful for improving the performance of MWP solving.


\subsection{Analysis of Data Augmentation}
Here we further investigate how the augmented data improves the MWP solving model.
In consideration of \textcolor{black}{the} balance of experiment time and accuracy,  we use the Prefix model on the development set of Math23K for analysis.







\begin{table}
\centering
\begin{tabular}{lll}
\hline

\bf Data & \bf   Data Prop.  & \bf Acc. \\ \hline \hline
Original only & --  & 68.0\%\\
RODA Only(2.24 times) & -- & 50.0\%\\
Orig+RODA & 1:0.5  & 69.5\% \\
Orig+RODA & 1:1 &  69.8 \% \\
Orig+RODA & 1:1.5 &  70.9 \%  \\
Orig+RODA(All) & 1:2.24 & \bf 71.0\% \\

\hline
\end{tabular}
\caption{Effects of data augmentation by adjusting the augmentation proportion on the devlopment set. The middle column denotes the proportion of original data and augmented data.}
\label{ratio}
\end{table}

Previous studies on data augmentation \cite{DBLP:conf/iclr/YuDLZ00L18} show that too much augmented data might harm the performance of the model. 
Thus, we experiment \textcolor{black}{with} what percentage of our augmented data can best improve the MWP solving model.
\textcolor{black}{We only use our augmented data to train the model and achieve the accuracy of 50\%, still far lower than only using the original Math23k training data (68\%).
The cause of this low performance could be caused by various reasons. First, the number identification stage filters out various types of questions, that the model can not deal with these unseen examples. Second, the new examples might introduce noise, that the augmented problem is similar to the original one in lexical terms and is hard to differ for the solver. Third, while our method can cover more mathematical knowledge points, it does not completely match with the original distribution. For example, considering the rectangle circumference formula $C=(a+b)×2$, the reversed formula $a=C/2 - b$ would appear less in the problem distribution. We leave these problems for future work.}

We also consider combining the original data with different proportions of augmentation data as training data whose performance is shown in Table \ref{ratio}.
We can see that, as the size of the augmented data increases, the performance stably increases, which shows how the size of the training data \textcolor{black}{affects} the performance. The model performs \textcolor{black}{similarly} when the proportion is 1.5 and using the whole augmented data. 
Because the performance of the model has not decreased along with the increase of augmented data, we use the full augmented data for the reported results. This can demonstrate the reverse operation can infer new knowledge points, which \textcolor{black}{helps to learn the mathematical reasoning logic} while more high-quality data can be used to well train the neural networks.







\begin{table}[t]
\centering
\begin{tabular}{llll}
\hline

\multicolumn{2}{l}{\bf Model} &  \bf Acc \\ \hline \hline
\multicolumn{2}{l}{BT Only} & 45.6 \% \\
\multicolumn{2}{l}{RODA(1:1) Only} & 48.2\%\\
\multicolumn{2}{l}{RODA(1:2.24) Only} & 50.0\%\\
\multicolumn{2}{l}{Origin} & 68.0\% \\
& + BT& 68.2 \%  \\
& + RODA(1:1) & 69.8\%\\
& + Full RODA & \bf 71.0\% \\

\hline
\end{tabular}
\caption{Comparison with Back Translate data augmentation method.}
\label{translate}
\end{table}

\begin{CJK*}{UTF8}{gbsn}
\begin{table*}[h]
\centering
\begin{tabular}{p{5pt}p{140pt}p{150pt}p{100pt}p{5pt}p{5pt}}
\hline

\bf \# & \bf Chinese Text & English Translation & Equation & Coh & Cor \\ \hline \hline
1& 甲,乙两同学相距$t_a$米,同时相向而行,乙同学速度为$t_b$米/秒,与此同时,一只小狗以$t_c$米/秒的速度,从甲身边跑向乙,遇到乙后又以同样的速度跑向甲,…如此往返,直到甲、乙同学相遇,问在此段时间内,小狗共跑了$t_d$米,甲同学速度为多少米/秒？&
The distance between A and B is $t_a$ meters. The two are heading toward each other. The speed of B is $t_b$ m/s. Meantime, a dog starts with A and then runs back and forth between the two people with a speed of $t_c$ m/s, until the two people meet. During this duration, the dog ran $t_d$ meters. What is the speed of A? &
$x = t_a * t_c / t_d - t_b$&
5 &1  \\
   
2 &几个 小朋友 分 苹果 , 如果 每人 分 $t_a$ 个 , 如果 每人 分 $t_b$ 个 , 少 $t_c$ 个 , 小朋友 有 $t_d$ 人 , 就 余 多少 个？&
A few children are sharing apples. If each child gets $t_a$ apples. If each child gets $t_b$ apples, there would be a shortage of $t_c$ apples. There are $t_d$ children. How many apples are left? &
$x = t_d * ( t_b - t_a ) - t_c$&
2& 0\\

\hline
\end{tabular}
\caption{Examples of the output from the Data Augmentation Module. \textbf{"Coh"} and \textbf{"Cor"} stands for Coherence and Correctness in Table \ref{human}.}
\label{ex}
\end{table*}
\end{CJK*}

\begin{CJK*}{UTF8}{gbsn}
\begin{table*}[h]
\centering
\begin{tabular}{p{120pt}|p{80pt}p{120pt}p{100pt}p{5pt}p{5pt}}
\hline

  \bf Original Text & \bf BT & \bf BT English & \bf Origin \& BT Equation & \bf Coh & \bf Cor \\ \hline 
某江汛期时\underline{中流}与\underline{沿岸}的水速有很大不同，中流每小时$t_a$里， 沿岸每小时$t_b$里。今有一汽船\underline{顺中流而下}， $t_c$小时行驶了$t_d$里，问从沿岸返回原处需几小时？& 在河流中间，中游和海岸的水流速度非常不同。中游时速为$t_a$英里/小时，海岸时速为每小时$t_b$英里。今天，有一艘汽船沿着河流航行，在$t_c$小时内行驶$t_d$英里。从海岸返回原始地点需要多长时间？& In the middle of a river, the water velocity of the middle stream and seacoast are very different. The middle stream velocity is $t_a$ miles per hour and the seacoast velocity speed is $t_b$ miles per hour. Today, a steamer sails down the river and travels for $t_d$ miles in $t_c$ hours. How long does it take to return to the starting point from the seacoast? & $x = t_d/(t_d/t_c-t_a-t_b)$ & 2 & 0\\ \hline
 \bf Origin Text English  & \bf RODA & \bf RODA English & \bf RODA Equation & \bf  N & \bf C \\ \hline 
 During the flood season of a certain river, the water velocity of the middle stream and the coastal waters are very different. The middle stream velocity is $t_a$ kilometers per hour and the coastal velocity speed is $t_b$ kilometers per hour. Now a steamer sails down the river in the midstream and travels for $t_d$ miles in $t_c$ hours. How many hours does it take to return to the starting point in the coastal waters?& 某江汛期时中流与沿岸的水速有很大不同 ，沿岸每小时$t_a$里 ，今有一汽船顺中流而下，$t_b$小时行驶了$t_c$里 ，问从沿岸返回原处需$t_d$小时 ，中流每小时几里？& During the flood season of a certain river, the water velocity of the midstream and the coastal waters are very different. The coastal velocity speed is $t_a$ kilometers per hour. Now a steamer sails down the river in the midstream and travels for $t_c$ miles in $t_b$ hours. If it takes $t_d$ to return to the starting point in the coastal waters, What is the midstream velocity? & $x = t_c/t_b - t_c/t_d - t_a$ & 5 & 1 \\
   

\hline
\end{tabular}
\caption{Comparison of Back Translation and Reversed Operation Based Data Augmentation. \textbf{"Coh"} and \textbf{"Cor"} stands for Coherence and Correctness in Table \ref{human}.}
\label{btex}
\end{table*}
\end{CJK*}

We also compare our data augmentation method with another novel data augmentation method, back-translate (BT) \cite{DBLP:conf/iclr/YuDLZ00L18} . For the BT method, we translate the problem text into English and then back to Chinese by Google Translate\footnote{https://translate.google.com/}. 
As BT can only \textcolor{black}{perform} data augmentation with the 1:1 proportion of the original data, we also control the size of our augmented data. Table \ref{translate} compares the MWP solving results.
As shown in Table \ref{translate}, 
We can see that the performance of BT 
has a significant gap with RODA, even when the data augmentation proportion is the same. There are two reasons for such \textcolor{black}{a} performance gap. \textcolor{black}{First,} BT may \textcolor{black}{introduce} more noise into the training data through the translation-based paraphrase and degrade the model performance, while our method can better ensure data quality. \textcolor{black}{Second,} BT only paraphrases the same meaning on the lexicon level, while our data augmentation method can introduce new knowledge points that \textcolor{black}{helps to learn the mathematical reasoning logic}.


   



\subsection{Human Evaluation}

To further examine the data quality of our augmented data, we perform human evaluation to examine the original data, back-translate augmented data, and RODA data.
We sample 100 MWPs from each dataset. 
\textcolor{black}{25 examples have two numbers, 25 examples have three, 25 examples have four, and the last 25 have five or more numbers. The sampling in each category is random.}
The evaluation involves two aspects. The first is the coherence which is ranked between 1-5. This examines whether the generated text is  coherent. The second is correctness which is classified \textcolor{black}{as} either 0 or 1. This examines whether the equation matches the problem text. Two annotators participate in ranking the data.
Table \ref{human} lists the average scores for each dataset with respect to the two metrics.
From Table \ref{human},  we can see our method can generate new data which has only a small performance gap with the original data. Compared to the BT method, our augmented data is \textcolor{black}{of} higher quality in both coherence and correctness. This can demonstrate how our method is reliable so that the key information is not lost, which is also why our augmented data can well boost the model performance. 

\begin{table}[h]
\centering
\begin{tabular}{lll}
\hline

\bf Model & \bf Coherence &  \bf Correctness \\ \hline \hline
Original& 4.27& 0.92 \\
BT & 3.24 & 0.55 \\
RODA & 3.86 & 0.84   \\

\hline
\end{tabular}
\caption{Human evaluation on Math23K.}
\label{human}
\end{table}


Here we show two augmented \textcolor{black}{examples} in Table \ref{ex}. In case 1, this newly produced example has coherent text and correct solution, even if the original mathematical logic and description text is fairly complex.
The original example involves the numbers that the time used by A, B, and the dog is the same so that it can form the problems for each of the 5 variables: the speed of A, B, and the dog, the distance between A and B and the distance that the dog has run. \textcolor{black}{In this example,} the original training data can  be augmented into 4 new \textcolor{black}{high-quality} question-answer pairs, and each of them holds a new \textcolor{black}{mathematical knowledge point}, which demonstrates the effectiveness of our model.  
We also show an example where our method failed in case 2. When swapping the order of the sentences, the coreference resolution of \textbf{apples} in the final question has changed that it no longer asks about the origin variable; \textcolor{black}{therefore,} the new reversed question would no longer be natural nor correct. In the case that the sentence order swapping would influence the coreference resolution of natural text, our method would no longer work. \textcolor{black}{Such error are caused since our question generation module does not consider the dependency between discourses. Such kind of error could be reduced with an additional discourse analysis module, which could be left for further work.}

We show one example of comparison between BT and RODA in Table \ref{btex}. As we can see, during the paraphrase BT translates \begin{CJK*}{UTF8}{gbsn}中流\end{CJK*} into \begin{CJK*}{UTF8}{gbsn}中游\end{CJK*} and \begin{CJK*}{UTF8}{gbsn}沿岸\end{CJK*} into \begin{CJK*}{UTF8}{gbsn}海岸\end{CJK*}, which makes the question no longer natural and reasonable. Meanwhile, it leaves out a key information point \begin{CJK*}{UTF8}{gbsn}顺中流而下\end{CJK*} that the question meaning has completely changed and the equation no longer matches with the question. Such kind of minor errors during the paraphrase would be fatal for MWP data augmentation that the key information is changed. The noisy new examples would reduce the performance of the model. Meanwhile, our method RODA correctly formed a natural reversed question with a new \textcolor{black}{mathematical knowledge point} corresponding to the reversed equation. 



   



\subsection{A Study on English Dataset}

\begin{table}[h]
\centering
\begin{tabular}{lll}
\hline

\multicolumn{2}{l}{\bf Model} &  \bf Acc \\ \hline \hline
\multicolumn{2}{l}{GTS} & 68.2\% \\
& + BT& 70.3 \%  \\
& + Full RODA & \bf 70.7\% \\

\hline
\end{tabular}
\caption{Results on AllArith.}
\label{en}
\end{table}

\begin{table}[t]
\centering
\begin{tabular}{l|ll}
\hline
Type & \# & Prop. \\
\hline
Original Problems & 831 & -\\
Filtered Problems& 1 & 0.00\\
Original Numbers & 1953 & 2.35\\
Candidate Numbers & 1951 & 2.35\\
Irreversible Numbers & 1236 & 1.49\\
Augmented Problems & 715 & 0.86\\
\hline
\end{tabular}
\caption{The statistics of the data augmentation on the full set of AllArith. \textbf{"Prop."} stands for the proportion of the item compared with original problems.
}
\label{tab:augen}
\end{table}
We also extend our data augmentation method to  AllArith\footnote{\textcolor{black}{Download link:https://github.com/CogComp/arithmetic}} \cite{roy2017unit}, which is a high-quality English MWP dataset with 831 problems, to show how our method works on an English dataset. As shown in Table \ref{tab:augen}, our data augmentation method generated 715 new examples for AllArith, respectively. We use the GTS model in this experiment and compare \textcolor{black}{it} with the BT data augmentation method.
The results are based on 5-cross validation following the split of \textcolor{black}{{Roy and Roth}~\cite{roy2017unit}}.
As we can see in Table \ref{en}, 
our model achieves performance improvement when adding the augmented data, which demonstrates the generalization ability of our method beyond language. Here the BT data also \textcolor{black}{achieve} performance improvement, slightly lower than using our augmented data. 
We consider the reason why our data augmentation method does not exhibit significant advantages over back translation for two reasons: First, the samples in Math23K have more variables in average (2.77 for Math23K and 2.35 for AllArith), that our method can generate more new augmented examples for Math23K; Second, the samples in Math23K are more difficult in mathematical reasoning complexity, our method can introduce new mathematical knowledge points that can benefit the model to learn mathematical reasoning logic. 
In the future, we will research how to produce more high-quality data.

\section{Conclusion}

In this paper, we propose the reverse operation based data augmentation for MWP solving, which converts the question and equation via reverse operation. Enlightened by how \textcolor{black}{humans} perform double-checking during calculation, 
the method can perform cheap and accurate data augmentation that could be adapted to any model.
The augmented data 
also provides supervision of new mathematical knowledge points that could benefit the model beyond paraphrasing the text. 
We evaluate our method on Math23K and achieve state-of-the-art performance.
\textcolor{black}{In} comparison with a strong baseline Back Translation, we show how our method significantly outperforms on a complex and large-scale dataset Math23k and achieve comparable performance on a simple and small dataset AllArith. 

\section*{Acknowledgments}
The authors would like to thank Haoran Zhang for his helpful advice for this research. This work was partially supported by National Key Research and Development Project (2019YFB1704002) and National Natural Science Foundation of China (61876009).

\appendices

\section{Removing Negative Terms}

\textcolor{black}{
To be noticed, sympy would put negative numbers in the first position of one equation during simplification, which causes the problem that one equation no longer can form a binary abstract syntax tree. 
We also design an algorithm to avoid this problem by moving the first addition term to the first position. We show an example in Table \ref{tab:equanorm}.}
\begin{table}[h]
\centering
\color{black}
\begin{tabular}{l|l}
\hline
Step & Equation\\
\hline
Equation&  $x=c-a-c+(c*a)+(b/b)$\\
Sympy & $x=-a+1+(a*c)$\\
Adjusted& $x=1 -a+(a*c)$\\
\hline
\end{tabular}

\caption{One example of equation normalization}
\label{tab:equanorm}
\end{table}

\textcolor{black}{
We show the algorithm for removing the negative terms in the front of the equation during equation normalization in Algorithm \ref{remove}. To be noticed, this process does not change the mathematical value of the equation but only how it is written.}

\floatname{algorithm}{\color{black}Algorithm}
\begin{algorithm}
\color{black}
\caption{\textcolor{black}{The algorithm of removing negative terms}}
\begin{algorithmic}[1]
\floatname{algorithm}{Procedure}
\renewcommand{\algorithmicrequire}{\textbf{Input:}}
\renewcommand{\algorithmicensure}{\textbf{Output:}}
\REQUIRE Original Equation
\ENSURE New Equation Written Form
\WHILE{True}
\IF{Brackets in Equation}
\STATE subs = list of sub-equation in the brackets;
 \FORALL{sub-equation in subs}
\STATE Remove Negative(sub-equation)
\ENDFOR
\ENDIF 
\IF{Equation[0]=='-'}
\STATE Find the first add token;
\STATE Move the token to the front;
\ENDIF
\ENDWHILE
\RETURN Equation
\end{algorithmic}
\label{remove}

\end{algorithm}

\section{Interrogative Pronouns}

\begin{CJK*}{UTF8}{gbsn}
\textcolor{black}{
We show the list of interrogative pronouns used for question conversion here in Table \ref{words}. Some of the interrogative pronouns also have other meanings in the declarative discourse unit, so we only detect them if they are in the last discourse unit of the sentence.}

\begin{table}[h]
\centering
\color{black}
\begin{tabular}{ll}
\hline

\bf Chinese & \bf English Translation   \\ \hline \hline
多少& How many/much \\
几分之几 & the fraction number is\\
几& How many  \\
=$\dag$ & =\\
求 & Please solve\\
((())/(())) & ((())/(()))\\
多$\dag$ & how much more than \\
\hline
\end{tabular}
\caption{A list of interrogative pronouns. $\dag$ denotes this pronoun is only valid when it is in the last discourse unit of the question.}
\label{words}
\end{table}

\end{CJK*}






\ifCLASSOPTIONcaptionsoff
  \newpage
\fi



\bibliographystyle{IEEEtran}
\bibliography{ieee}
%

%

\begin{IEEEbiography}[{\includegraphics[width=1in,height=1.25in,clip,keepaspectratio]{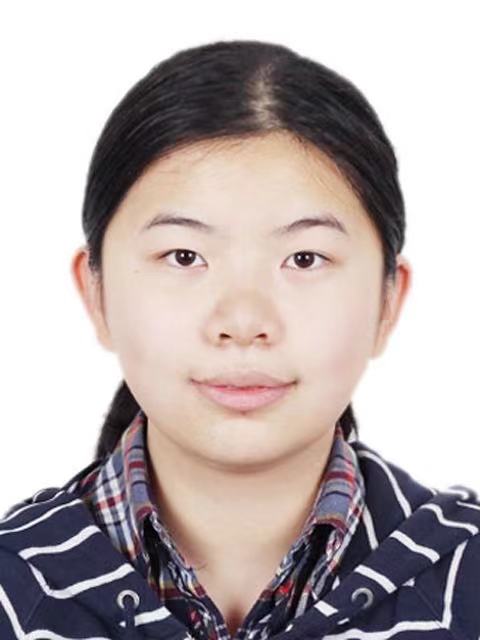}}]{Qianying Liu}
Qianying Liu received her B.S. in Information and Computing Science from Peking University, China in 2018 and M.S. in Intelligence Science and Technology from Kyoto University, Japan in 2020, respectively. She is currently an Ph.D. candidate and research assistant at Kyoto University. Her major research interests include information extraction, text generation and question answering.
\end{IEEEbiography}

\begin{IEEEbiography}[{\includegraphics[width=1in,height=1.25in,clip,keepaspectratio]{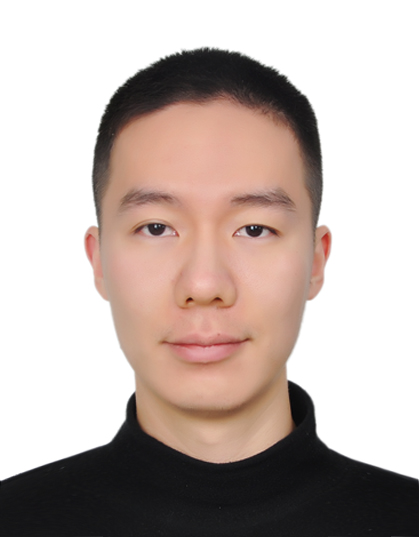}}]{Wenyu Guan}
Wenyu Guan received the BSc degree from the Xian Jiaotong University and MSc degree from Peking University, China, in 2017 and 2020, respectively. He worked as a  software engineer in the Institute of Information Engineering - Chinese Academy of Science from 2017 to 2019.  He served as an intern research assistant in the Institute of Computational Linguistics, Peking University from 2019 to 2020. His research interest includes math word problem solving, semantic parsing, and information retrieval.
\end{IEEEbiography}

\begin{IEEEbiography}[{\includegraphics[width=1in,height=1.25in,clip,keepaspectratio]{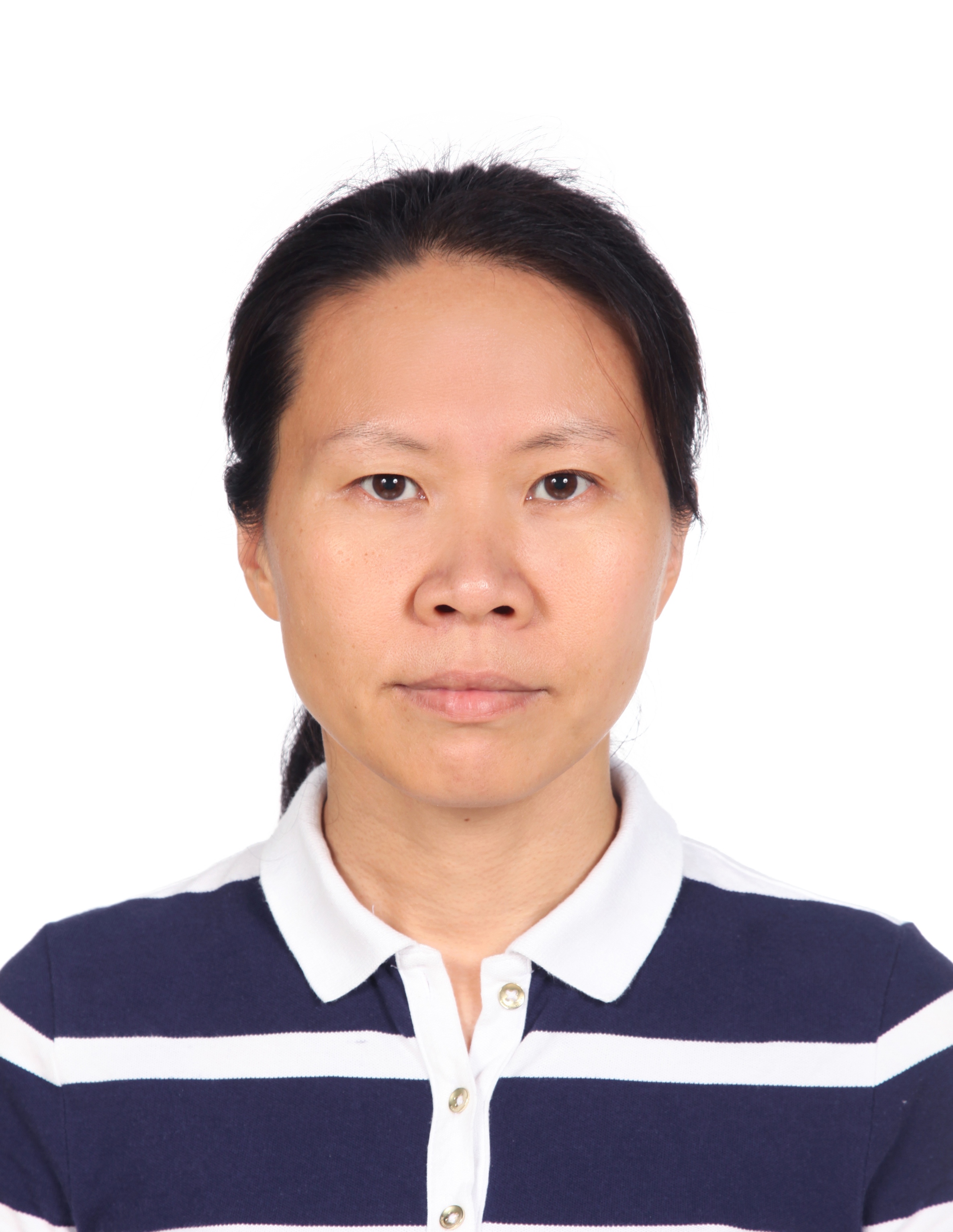}}]{Sujian Li}
Sujian Li received the Ph.D. degree from the Institute of Computing Technology, Chinese Acamedy of Sciences, Beijing, China, in 2002. She is currently a tenured Associate Professor at the Institute of Computational Linguistics, Peking University, Beijing. Her main research interests include natural language processing, information extraction, and document summarization.
\end{IEEEbiography}

\begin{IEEEbiography}[{\includegraphics[width=1in,height=1.25in,clip,keepaspectratio]{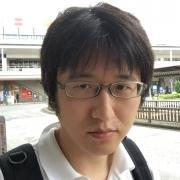}}]{Fei Cheng}
Fei Cheng received his B.S. in Applied Physics from Donghua University in 2005 and M.S. \& Ph.D. in Informatics from NARA Institute of Science and Technology in 2013 and 2018. His research interests include morphological analysis, information extraction, and a broad range of natural language processing topics. Currently, he is a program-specific assistant professor at Kyoto University.
\end{IEEEbiography}

\begin{IEEEbiography}[{\includegraphics[width=1in,height=1.25in,clip,keepaspectratio]{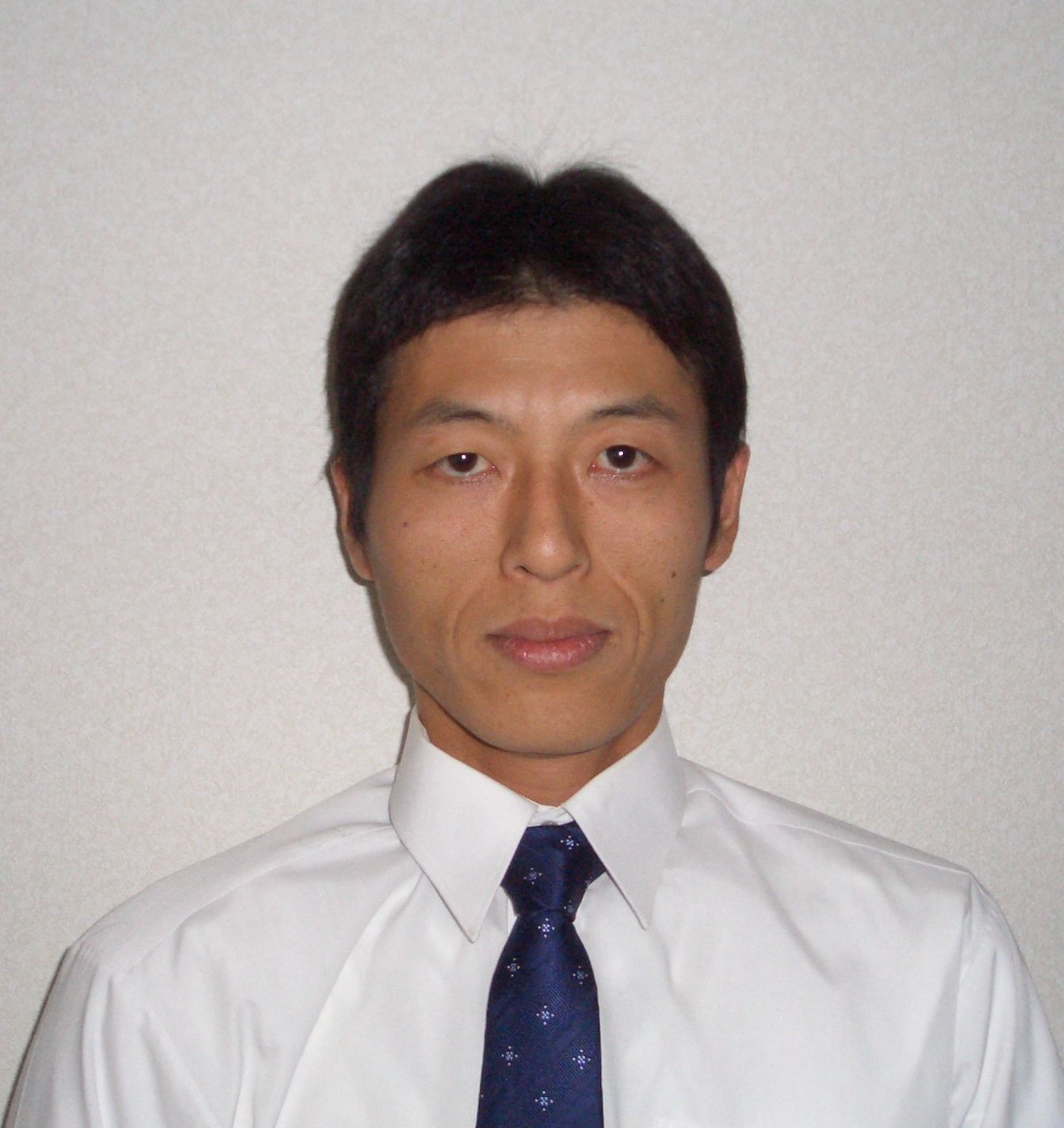}}]{Daisuke Kawahara}
Daisuke Kawahara received his B.S. and M.S. in Electronic Science and
Engineering from Kyoto University in 1997 and 1999, respectively. He
obtained his Ph.D. in Informatics from Kyoto University in 2005. He is
currently a professor of the department of communications and computer
engineering, Waseda University. His research interests center on
natural language processing, particularly knowledge acquisition and
text understanding.
\end{IEEEbiography}

\begin{IEEEbiography}[{\includegraphics[width=1in,height=1.25in,clip,keepaspectratio]{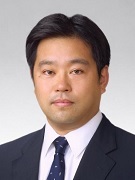}}]{Sadao Kurohashi}
Sadao Kurohashi received the B.S., M.S., and PhD in Electrical
Engineering from Kyoto University in 1989, 1991 and 1994,
respectively.  He has been a visiting researcher of IRCS, University
of Pennsylvania in 1994.  He is currently a professor of the Graduate
School of Informatics at Kyoto University.
\end{IEEEbiography}







\end{document}